\documentclass[conf]{new-aiaa}
\usepackage[utf8]{inputenc}
\usepackage{mathtools}
\usepackage{graphicx}
\usepackage{amsmath}
\usepackage[version=4]{mhchem}
\usepackage{siunitx}
\usepackage{booktabs}
\usepackage{longtable,tabularx}
\usepackage{todonotes}
\usepackage{caption}
\usepackage{subcaption}

\title{High Cycle S-N curve prediction for Al 7075-T6 alloy using Recurrent Neural Networks (RNNs)}

\author{Aryan Patel}
\affil{Dept. of Mechanical Engineering, Stanford University}
\affil{apatel72@stanford.edu}

\begin{document}

\maketitle
\begin{abstract}
Aluminum is a widely used alloy, which is susceptible to fatigue failure. Characterizing fatigue performance for materials is extremely time and cost demanding, especially for high cycle data. To help mitigate this, a transfer learning based framework has been developed using Long short-term memory networks (LSTMs) in which a source LSTM model is trained based on pure axial fatigue data for Aluminum 7075-T6 alloy which is then transferred to predict high cycle torsional S-N curves. The framework was able to accurately predict Al torsional S-N curves for a much higher cycle range. It is the belief that this framework will help to drastically mitigate the cost of gathering fatigue characteristics for different materials and help prioritize tests with better cost and time constraints.
\end{abstract}
\section{Introduction}
Aluminum and its alloys are widely used in multiple industries, such as construction, aerospace, and automotive due to their light weight and advantageous material properties. However, one of the major limitations of introducing lightweight materials like aluminum is the high cost to characterize its mechanical performance. Fatigue accounts for 80 percent of engineering failure of designed metal parts \cite{bobbyj}. Unlike Steel and its alloys which possess characteristics for infinite fatigue life, Aluminum alloys will always fail under fatigue loads after a certain number of cycles. Hence, it becomes important to be able to characterize fatigue strength of Aluminum alloys. Generally, fatigue life is predicted utilizing the Wohler $S-N$ fatigue curve which express the lifetime of the material as measured by the logarithm of the number of cycles $N$, plotted against the logarithm of $S$ \cite{bobbyj}. These curves are generated using experimental test data, which are incredibly expensive and time consuming to obtain. In addition, random variations in the test procedure may affect results. The underlying mechanisms behind fatigue failure are also non-linear \cite{Al_ANN_prob}. All of these factors combined lead to utilizing data driven and Machine Learning methods to predict fatigue life. 

There have been several avenues to apply ML techniques to the problem of fatigue life prediction. Chen et al. compiled a comprehensive review on Neural Net techniques used in Fatigue research \cite{fatigue_nn_review}. This review goes over techniques and provides relevant articles from simple Feed Forward Neural Networks to Recurrent Neural Networks (RNNs) as well as Physics informed Neural Networks (PINNs). Martinez et al. proposed the use of Artificial Neural Networks with Bayesian Regularization to predict Fatigue life for Aluminum \cite{Al_ANN_prob}. Durodola et al. developed an Artificial Neural Network for fatigue analysis including the effects of mean stress \cite{DURODOLA2018321}. Wei et al. proposed the use of LSTMs coupled with Transfer Learning to predict high cycle fatigue performance using low cycle data \cite{WEI2022107050}. The dataset chosen for training and testing has been selected from the experimental fatigue data collected by Kluger et al. \cite{datasheet}. Due to the comparatively low number of samples present in the dataset, synthetic data has been generated as described in \cite{WEI2022107050} \cite{YANG2021121761}. The final dataset contains two curves, one characterizing fatigue strength under applied rotating bending axial alternating loads and the other dealing with applied torsional loads under reversed torsion tests.

This paper attempts to characterize the high cycle fatigue strength of 7075-T6 Al alloy using primarily low cycle data to make predictions for fatigue strength at high cycles. Although the S-N curve for Aluminum 7075 is monotonically decreasing, the overall trend of the curve changes with the number of cycles. As this is inherently a prediction problem involving extrapolation from existing data, standard Deep Neural Networks (DNNs) are not optimal. For this reason, Long Short-term Memory Networks (LSTMs) are chosen to deal with this problem. LSTMs are a type of Recurrent Neural Networks (RNNs) useful for sequence learning tasks \cite{LSTM}. In recent years, LSTMs have been utilized to predict the life of Li-ion batteries \cite{LSTM_battery} as well predict the stress-strain curve of materials under uni-axial compression \cite{LSTM_stress}. In this study, LSTMs have been utilized to predict fatigue strength at high cycles from data taken under high stress amplitudes at low cycles.  In addition, to properly utilize the available low cycle data, a transfer learning approach has also been used as described in \cite{TR} \cite{TR_LSTM}. 

In summary, a TR-LSTM framework has been developed, wherein the reversed torsion S-N curves prediction of 7075-T6 Al was transferred from corresponding rotating bending S-N data. In the framework, based on the source LSTM predictive models for rotating bending curves, the later stages of reversed torsion S-N curves were predicted by TR-LSTM models.

\section{Methods}

\subsection{Dataset and Data Processing}
The dataset utilized here is obtained from experimental fatigue data collected by Kluger et al. \cite{datasheet}. The data here contains experimental data for rotating bending and reversed torsion fatigue tests for Al 7075-T6 alloy along with fitted curves according to the formula as per \cite{Sakai} :

    \begin{equation}
        \sigma = 10^{Alog10(N)+B} + D
    \end{equation}

The curve contains data for equally 1000 log-spaced points for number of cycles from $5e3$ to $3e6$ for both the axial and torsion data. This data was split into training and test sets. For the rotating bending curves, the data before the cycle of $2.31e5$  (containing 600 data points) were selected as the training set for the source LSTM model; the remaining data at lower stress amplitudes were used for model testing. For the prediction of the reversed torsion curves, only the data before the cycle of $3.4e4$ (containing 300 data points) were used for training the TR-LSTM model.

\subsection{TR-LSTM framework}
\begin{figure}[h!]
        \centering
        \includegraphics[width=\textwidth*2/3]{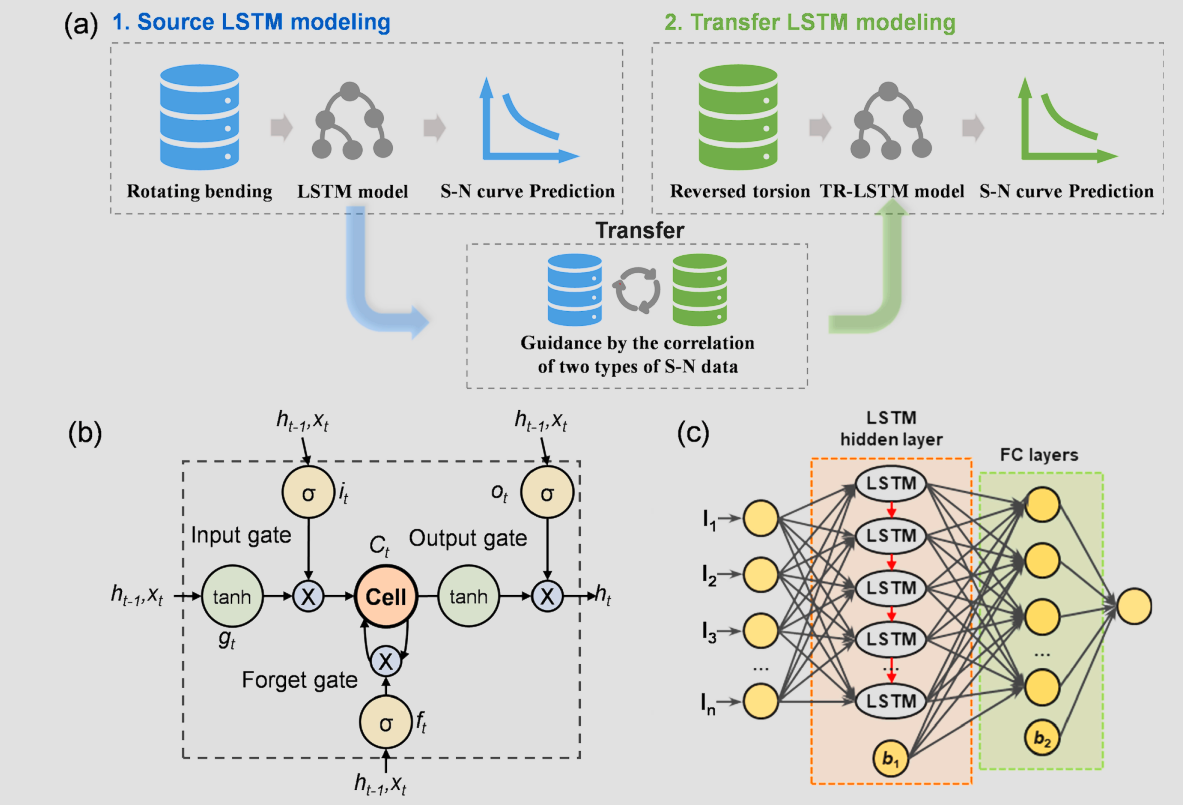}
        \caption{Schematic of TR-LSTM for S-N curve: (a) the framework of transfer learning in this research; (b) LSTM cell; (c) the architecture of LSTM model. \cite{WEI2022107050}}
        \label{fig:LSTM_arch}
\end{figure}

The LSTM layer is the basic of TR-LSTM, and the details of the LSTM layer are first enumerated and described below. The architecture of LSTM cell is schematically shown in Figure 1(b). During training of LSTM, the input, output, and forget gates allow the LSTM to forget or write new information to the memory cell. $x_t$ is the input layer at the moment $t$. $h_t$ is the hidden layer at the moment $t$, representing short-term state, which is also the output of the LSTM cell. $f_t$, $i_t$, $o_t$ are the forget, input and output gates, respectively. $C_t$ is the long-term cell state. $g_t$ is input node $tanh$ layer, which generates a vector of the new candidate state. The equations to calculate these variables are as follows: \\
Forget gate:
\begin{equation}
    f_t = \sigma(W_{fx}x_t + W_{fh}h_t - 1 + b_f)
\end{equation}
Input gate:
\begin{equation}
    i_t = \sigma(W_{ix}x_t + W_{ih}h_t - 1 + b_i)
\end{equation}
\begin{equation}
    g_t = tanh(W_{gx}x_t + W_{gh}h_t - 1 + b_g)
\end{equation}
Long-term cell state is updated as:
\begin{equation}
    C_t = g_t*i_t + C_{t-1}*ft
\end{equation}
Output gate:
\begin{equation}
    o_t = \sigma(W_{ox}x_t + W_{oh}h_t - 1 + b_o)
\end{equation}
\begin{equation}
    h_t = o_t*tanh(C_t)
\end{equation}
where $W$ and $b$ values are the weights and biases, respectively. $\sigma$ and $tanh$ are sigmoid and hyperbolic tangent activation functions, respectively. \autoref{fig:LSTM_arch} shows a schematic of the proposed model. 
To get the final predicted stress output, the LSTM layers were connected to a fully connected $FC$ layer which was connected to the output. The number of hidden neural units in the LSTM layer and the number of neural units in the FC layer were both set to 64. Then, the target TR-LSTM models for the reversed torsion S-N curves prediction were trained as follows: (1) LSTM hidden layers of the source model were copied to the corresponding layers of the target model, called transferred layers. The transferred feature layers remained frozen and did not participate in further training. (2) The adjustable FC layer of the target model was then randomly initialized and trained for the reversed torsion S-N curves. 

For comparison's sake, the torsional data was used to train a non-TR LSTM as a reference model. In addition, both datasets were used to train a standard DNN with 4 hidden layers with 32 hidden units and a $tanh$ acitvation function to showcase the superiority of LSTMs in prediction compared to standard DNNs. 

The LSTM was trained based on a sequence length of 50, meaning that the training data was divided up into 50 different points and the 51st point was predicted using the prior 50. The source LSTM model was trained using the axial training data and then the TR-LSTM model for the torsional data was trained. For prediction of the testing data, stress amplitude at a certain cycle is obtained according to a sequence of predicted stress amplitudes prior to that cycle. Specifically, the first stress amplitude in testing data is predicted based on the tail of training data with the length of 50. The second stress amplitude is predicted based on a portion of the training set data data tail (length is 49) and the predicted first stress amplitude in testing data. And so on, the 51st stress amplitude is predicted based on first fifty predicted stress amplitudes in testing data. All LSTM models were trained for 500 epochs.

Root Mean Squared Error (RMSE) was used as the metric to judge predictive performance, given by:
\begin{equation}
    RMSE = (\frac{1}{n} \sum_{i=1}^{n} (f(x_i) - y_i)^2)^{\frac{1}{2}}
\end{equation}

\section{Results}

\subsection{Axial Data}
\begin{figure*}[t!]
    \centering
    \begin{subfigure}[t]{0.5\textwidth}
        \centering
        \includegraphics[height=2.in]{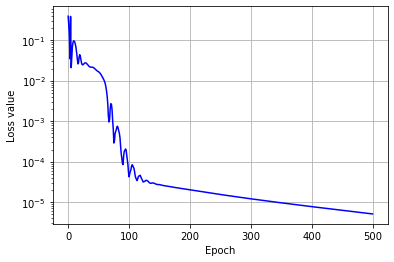}
        \caption{LSTM Training loss curve for Axial data}
        \label{fig:axial_loss}
    \end{subfigure}%
    ~ 
    \begin{subfigure}[t]{0.5\textwidth}
        \centering
        \includegraphics[height=2.in]{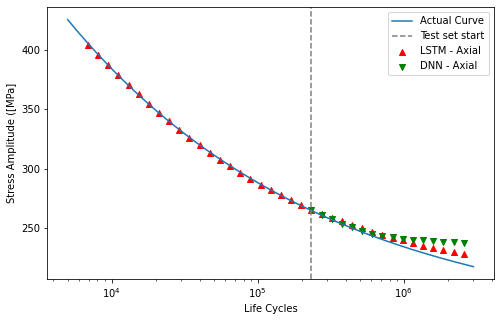}
        \caption{Axial S-N curve}
        \label{fig:axial_res}
    \end{subfigure}
    \caption{Prediction results for Axial data}
\end{figure*}

\autoref{fig:axial_loss} shows the training losses obtained from training the source LSTM. \autoref{fig:axial_res} shows the predicted results of the trained source LSTM model for the Axial data. It can be seen from the plot that the Source LSTM does a better job at predicting the data compared to the DNN, with a test RMSE error is 27.63 [MPa]. However, the LSTM deviates from the actual curve towards the end of the test cycle. A possible reason is the model being too simple to capture the trend based off the training data. Increasing the number of hidden units in the LSTM layer may help mitigate this.

\subsection{Torsional Data}
\begin{figure*}[t!]
    \centering
    \begin{subfigure}[t]{0.5\textwidth}
        \centering
        \includegraphics[height=2.in]{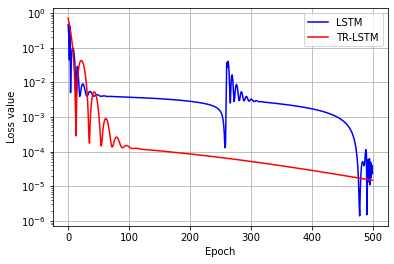}
        \caption{LSTM Training loss curve for Torsional data}
        \label{fig:tors_loss}
    \end{subfigure}%
    ~ 
    \begin{subfigure}[t]{0.5\textwidth}
        \centering
        \includegraphics[height=2.in]{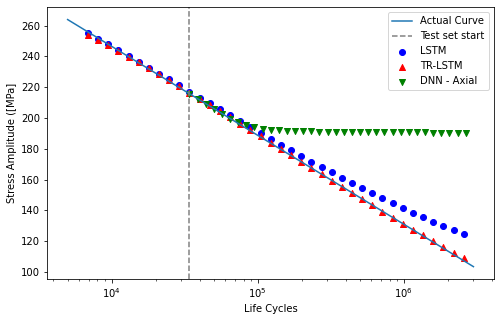}
        \caption{Torsional S-N curve}
        \label{fig:tors_res}
    \end{subfigure}
    \caption{Prediction results for Torsional Data}
\end{figure*}

\autoref{fig:tors_loss} shows the training losses obtained from training both the TR-LSTM and the standard LSTM on the torsional data. \autoref{fig:tors_res} shows the predicted results. Here, the training dataset is smaller compared to the axial data. As a result, it can be observed that the standard LSTM diverges from the actual curve at high cycles. The DNN perfoms the worst at extrapolation and prediction, as expected. The transfer learning based TR-LSTM is able to accurately predict high cycle torsional fatigue strength. This is evident from the test data RMSE values. The TR-LSTM had a test RMSE = 0.53 MPa while the standard LSTM had a test RMSE = 71.39 [MPa]. 

\section{Conclusion}
In conclusion, torsional fatigue S-N curves were predicted using transfer learning TR-LSTMs where the LSTM layers were transferred from a model trained with purely axial alternating stresses. The TR-LSTM model was able to accurately predict the fatigue strength at high cycles. However, further work must be performed to determine the generality of the model, with different Aluminum alloys and further materials. These studies require the need for additional data unavailable at this time. In addition, the LSTM is inherently Non-Markovian and requires the storage of previous results to make a prediction. Another avenue to explore is the usage of a Markovian model which can predict the future fatigue curve based primarily on an intial state.



\bibliography{sample}

\end{document}